\documentclass[conference]{IEEEtran}
\IEEEoverridecommandlockouts
\usepackage{cite}
\usepackage{amsmath,amssymb,amsfonts}
\usepackage{algorithmic}
\usepackage{graphicx}
\usepackage{textcomp}
\usepackage{xcolor}
\usepackage{CJKutf8}
\usepackage{subcaption}

\def\BibTeX{{\rm B\kern-.05em{\sc i\kern-.025em b}\kern-.08em
		T\kern-.1667em\lower.7ex\hbox{E}\kern-.125emX}}

\begin{document}
\begin{CJK*}{UTF8}{gbsn}
	
	\title{Enhancing Essay Scoring with Adversarial Weights Perturbation and Metric-specific AttentionPooling \\ 
	}
	
        \author{
            \IEEEauthorblockN{Jiaxin Huang}
            \IEEEauthorblockA{\textit{Information Studies} \\
                \textit{Trine University}\\
                Phoenix, USA \\
                jiaxinhuang1013@gmail.com}
            \and
            \IEEEauthorblockN{Xinyu Zhao\textsuperscript{*}}
            \IEEEauthorblockA{\textit{Information Studies} \\
                \textit{Trine University}\\
                Phoenix, USA \\
                lution798@gmail.com}
            \and
            \IEEEauthorblockN{Chang Che\textsuperscript{*}}
            \IEEEauthorblockA{\textit{Mechanical engineering} \\
                \textit{The George Washington University}\\
                Atlanta, USA \\
                liamche1123@outlook.com}
            \and
            \IEEEauthorblockN{Qunwei Lin\textsuperscript{*}}
            \IEEEauthorblockA{\textit{Information Studies} \\
                \textit{Trine University}\\
                Phoenix, USA \\
                linqunwei1030@outlook.com}
            \and
            \IEEEauthorblockN{Bo Liu\textsuperscript{*}}
            \IEEEauthorblockA{\textit{Software Engineering} \\
                \textit{Zhejiang University} \\
                Shanghai, China \\
                lubyliu45@gmail.com}
        }

	\maketitle
	
	\begin{abstract}
        The objective of this study is to improve automated feedback tools designed for English Language Learners (ELLs) through the utilization of data science techniques encompassing machine learning, natural language processing, and educational data analytics. Automated essay scoring (AES) research has made strides in evaluating written essays, but it often overlooks the specific needs of English Language Learners (ELLs) in language development. This study explores the application of BERT-related techniques to enhance the assessment of ELLs' writing proficiency within AES. 

        To address the specific needs of ELLs, we propose the use of DeBERTa, a state-of-the-art neural language model, for improving automated feedback tools. DeBERTa, pretrained on large text corpora using self-supervised learning, learns universal language representations adaptable to various natural language understanding tasks. The model incorporates several innovative techniques, including adversarial training through Adversarial Weights Perturbation (AWP) and Metric-specific AttentionPooling (6 kinds of AP) for each label in the competition.

        The primary focus of this research is to investigate the impact of hyperparameters, particularly the adversarial learning rate, on the performance of the model. By fine-tuning the hyperparameter tuning process, including the influence of 6AP and AWP, the resulting models can provide more accurate evaluations of language proficiency and support tailored learning tasks for ELLs. This work has the potential to significantly benefit ELLs by improving their English language proficiency and facilitating their educational journey.
	\end{abstract}
	
	
	\section{Introduction}
    
        Automated essay scoring (AES)\cite{dikli2006overview}\cite{lin2003automatic}\cite{brent2006automated} research has made significant strides in developing tools to evaluate and provide feedback on written essays. Moreover, AES research advances with course activities, NLP, and bi-modal approaches\cite{ashenafi2015predicting}\cite{rokade2018automated}\cite{gao2021bi}. However, these systems often fail to address the specific needs of English Language Learners (ELLs), who face unique challenges in language development. To bridge this gap, it is essential to define the ELLs within the context of AES research and explore the application of BERT-related methods to improve the assessment of ELLs' writing proficiency. 

        One promising avenue for enhancing AES is the exploration of BERT-related techniques\cite{mayfield2020should}\cite{beseiso2020empirical}\cite{xue2021hierarchical}. The landscape of natural language processing tasks has been transformed by BERT (Bidirectional Encoder Representations from Transformers), a language model based on the transformer architecture.\cite{devlin2018bert}. Several studies have investigated the effectiveness of BERT and its variants in AES, providing valuable insights into their impact on scoring accuracy and efficiency. However, the unique characteristics of ELLs require further attention in this context.

        To address the specific needs of ELLs, we propose the use of DeBERTa\cite{he2021debertav3}, a cutting-edge neural language model, to enhance automated feedback tools for ELLs. DeBERTa integrates state-of-the-art techniques, including adversarial training using Adversarial Weights Perturbation (AWP)\cite{wu2020adversarial}, and Metric-specific AttentionPooling (6AP) for each label in the competition. By leveraging these innovations, our proposed approach aims to improve the accuracy and effectiveness of automated feedback systems for ELLs, facilitating their language development and academic progress.

        The main objective of this research is to analyze the influence of hyperparameters, specifically the adversarial learning rate ($\text{adv\_lr}$) and adversarial perturbation magnitude ($\text{adv\_eps}$), on the performance of the DeBERTa model. The research aims to optimize the hyperparameter tuning process to enhance the model's ability to evaluate the language proficiency of English Language Learners (ELLs). By carefully adjusting these hyperparameters, we can fine-tune the model to achieve optimal results in assessing the writing skills of ELLs. The findings of this study will contribute to improving the accuracy and effectiveness of automated feedback tools for ELLs, supporting their language development and academic progress.

        By sensitizing automated feedback tools to the language proficiency of ELLs using DeBERTa with the added Metric-specific AttentionPooling (6 kinds of AP)\cite{er2016attention}\cite{li2018attention} technique, the resulting models can provide more accurate evaluations and enable teachers to assign suitable learning tasks for language development. This research endeavors to leverage data science techniques to create impactful solutions that support the educational journey of ELLs and contribute to their overall academic success.

	\section{RELATED WORK}
        Several early papers laid the foundation for automated essay scoring (AES) research. Dikli \cite{dikli2006overview} conducted a comparative study of automated scoring models for advanced English as a second language essays, highlighting the need for effective evaluation approaches. Lin and Hovy \cite{lin2003automatic}  focused on automatic evaluation of summaries using N-gram co-occurrence statistics, which demonstrated potential for assessing summary quality. Brent and Townsend \cite{brent2006automated} investigated automated essay grading in the sociology classroom, aiming to find common ground between automated systems and human graders. 

        Advancements in AES research highlight the utilization of course activities, NLP techniques, and bi-modal approaches for academic achievement prediction and automated grading. Ashenafi et al.\cite{ashenafi2015predicting} propose a predictive model that utilizes students' course activities to forecast their final exam scores. Their research highlights the potential of leveraging students' engagement and performance in course-related activities to assess their overall academic achievement. Rokade et al.\cite{rokade2018automated} present an automated grading system using natural language processing techniques. Their system analyzes and evaluates students' written responses, providing automated and efficient grading. Gao et al.\cite{gao2021bi} introduce a bi-modal automated essay scoring system designed specifically for handwritten essays. By considering both visual and textual information, their model demonstrates improved performance in scoring handwritten essays.

        Some studies contribute to the advancement of AES by investigating BERT-related techniques and their impact on scoring accuracy and efficiency. Mayfield and Black \cite{mayfield2020should} investigate the fine-tuning of BERT for automated essay scoring (AES), exploring different strategies and domain-specific training data. Their study provides insights into the effectiveness of BERT fine-tuning for AES and the impact of various factors on performance. Beseiso and Alzahrani\cite{beseiso2020empirical} empirically analyze BERT embeddings for AES, evaluating their quality and influence on scoring accuracy. Their research sheds light on the strengths and limitations of BERT embeddings in the context of essay scoring. Xue et al.\cite{xue2021hierarchical} propose a hierarchical BERT-based transfer learning approach for multi-dimensional essay scoring, leveraging the structure of essays to improve performance. 
        
        The integration of BERT and its variants, such as DebertaV3, has revolutionized automated essay scoring. Devlin et al.\cite{devlin2018bert} introduced BERT, which captures intricate language patterns and enhances evaluation accuracy. He et al.\cite{he2021debertav3} proposed DebertaV3, incorporating electra-style pre-training and gradient-disentangled embedding sharing, resulting in improved performance and efficiency. Wu et al.\cite{wu2020adversarial} explored adversarial weight perturbation to enhance robust generalization. These advancements highlight the potential of advanced pre-training, fine-tuning, and perturbation techniques in advancing automated essay scoring. Overall, these studies contribute valuable insights to the ongoing development and improvement of automated essay scoring systems.
		
	\section{ALGORITHM AND MODEL}
  
            \subsection{Notation} 
        In this study, we propose the use of DeBERTa, a state-of-the-art neural language model, to enhance automated feedback tools for English Language Learners (ELLs) in the context of automated essay scoring (AES). DeBERTa incorporates several innovative techniques, including adversarial training through Adversarial Weights Perturbation (AWP), and Metric-specific AttentionPooling (6 kinds of AP) for each label in the competition.

        Figure 1 illustrates the architecture of the DeBERTa model. The model consists of multiple layers of self-attention and feed-forward neural networks, allowing it to capture intricate language patterns and representations. 

                \begin{figure}[htbp]
                  \centering
                  \includegraphics[width=1\linewidth, height=4cm]{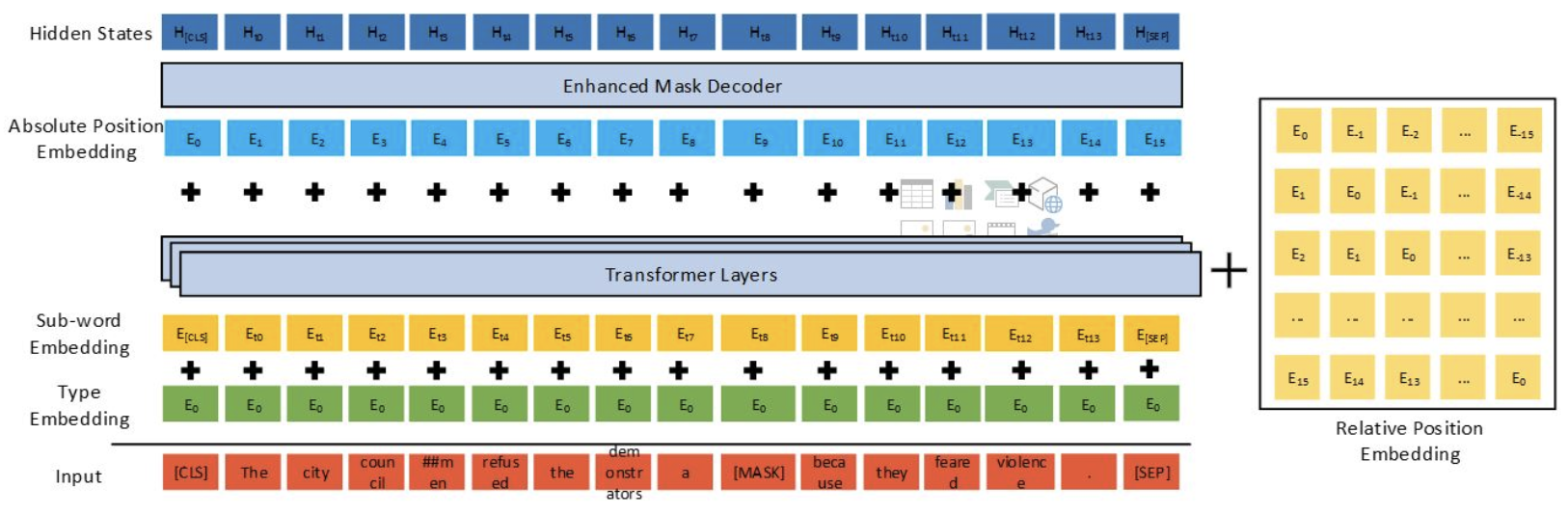}
                  \caption{DeBERTa model}
                  \label{fig:DeBERTa model}
                \end{figure}

        By leveraging these techniques, DeBERTa enables more accurate and efficient evaluation of ELLs' writing proficiency. This empowers automated feedback tools to provide targeted and personalized assessments for ELLs, supporting their language development and academic success in AES.

        Our model's performance and generalization ability were evaluated using a 5-fold Cross-Validation\cite{browne2000cross} strategy, incorporating MultiLabelStratifiedKFold to ensure fair label distribution. This is particularly relevant for the automated essay scoring (AES) task, which focuses on supporting English Language Learners (ELLs) in their language development. By enhancing automated feedback tools and improving scoring accuracy, we aim to provide personalized assessments for ELLs. The competition's evaluation metric, MCRMSE (Mean Columnwise Root Mean Squared Error)\cite{chai2014root}\cite{guo2005mutual}, calculates the RMSE for each label column and averages the results. This widely-used metric captures the differences between predicted and true values, providing an overall assessment of the model's performance across all labels.
        
            \subsection{Adversarial Weights Perturbation}

        In this study, our primary research objective is to investigate the impact of hyperparameters, particularly the adversarial learning rate ($\text{adv\_lr}$) and adversarial perturbation magnitude ($\text{adv\_eps}$), on the performance of the DeBERTa model. Through meticulous hyperparameter tuning, our aim is to achieve optimal performance in assessing the language proficiency of English Language Learners (ELLs).

        The inclusion of adversarial training has significantly improved our cross-validation (CV) score. While hyperparameters play a crucial role. To illustrate the Adversarial Weights Perturbation (AWP), we provide the following figure 2. Please note that our experiences may deviate slightly from the expected behavior, and there may be room for error in our understanding.

        \begin{figure}[htbp]
        \centering
        \includegraphics[width=1\linewidth]{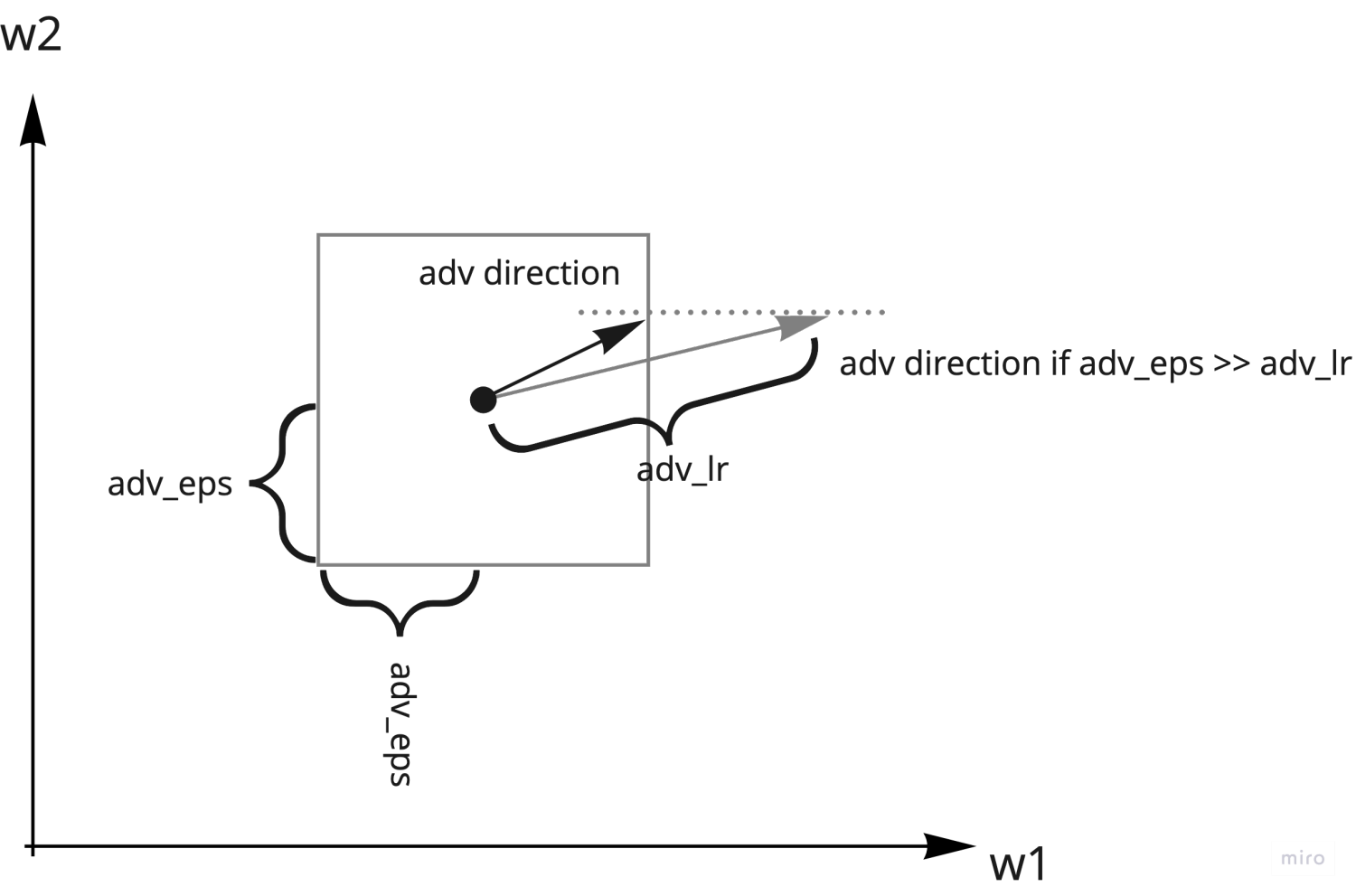}
        \caption{Illustration of the Adversarial Weights Perturbation (AWP)}
        \label{fig:AWP}
        \end{figure}

        By initiating the Adversarial Weights Perturbation (AWP) technique from epoch 2, our model is able to explore a broader range of optimization scenarios, thereby improving its generalization capability. AWP introduces smoothness to the loss function, preventing the model from getting stuck in local optima and facilitating convergence towards global optima. As a result, we observe a significant reduction in loss and improved model convergence. Moreover, the enhanced generalization ability of AWP leads to higher performance on the validation set, resulting in better validation scores.

        To provide a visual representation of this training process, we have included the following figure 4:

        \begin{figure}[htbp]
        \centering
        \includegraphics[width=1\linewidth]{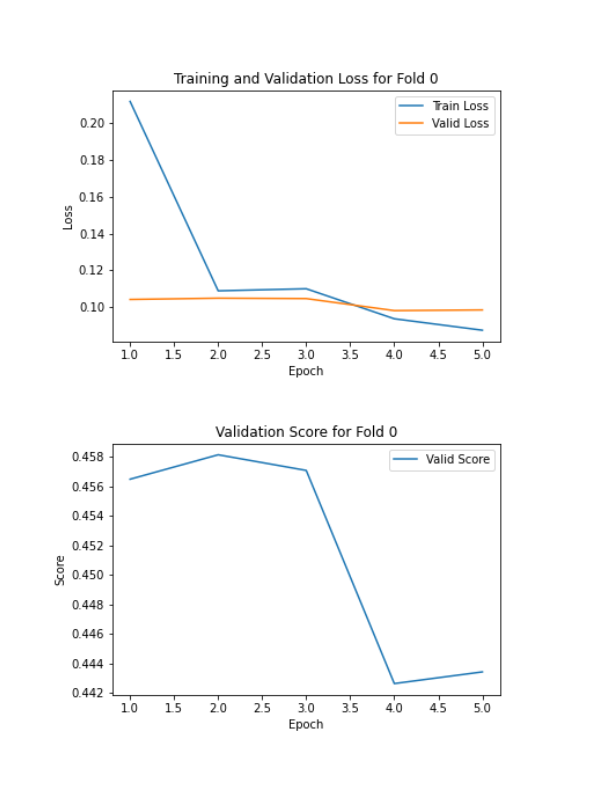}
        \caption{AWP Training Process}
        \label{fig:awp_training_process}
        \end{figure}

        This figure illustrates how AWP promotes smoother optimization and helps the model navigate through various training stages, ultimately leading to improved performance. The AWP technique plays a crucial role in enhancing the training dynamics and overall effectiveness of our model.

            \subsection{Metric-specific AttentionPooling}

        The Metric-specific AttentionPooling (6 kinds of AP) technique enhances the interpretability and performance of our automated essay scoring (AES) system by allowing us to assign different attention weights to different parts of the essays based on the specific metric being assessed. For example, when evaluating the coherence or organization of the essay, the model can prioritize attention on sentence-level or paragraph-level information. Conversely, when assessing vocabulary or grammar, the model can allocate attention to individual words or phrases.

        To visually illustrate the concept of 6 kinds of AP, Figure 3 depicts the dynamic attention allocation process within our model.

        \begin{figure}[htbp]
            \centering
            \begin{subfigure}{\linewidth}
                \centering
                \includegraphics[width=\linewidth]{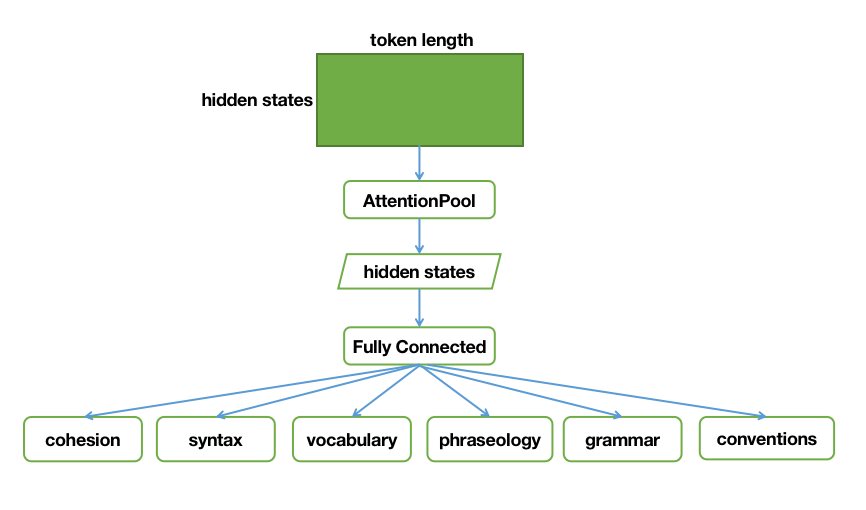}
            \end{subfigure}
            \begin{subfigure}{\linewidth}
                \centering
                \includegraphics[width=\linewidth]{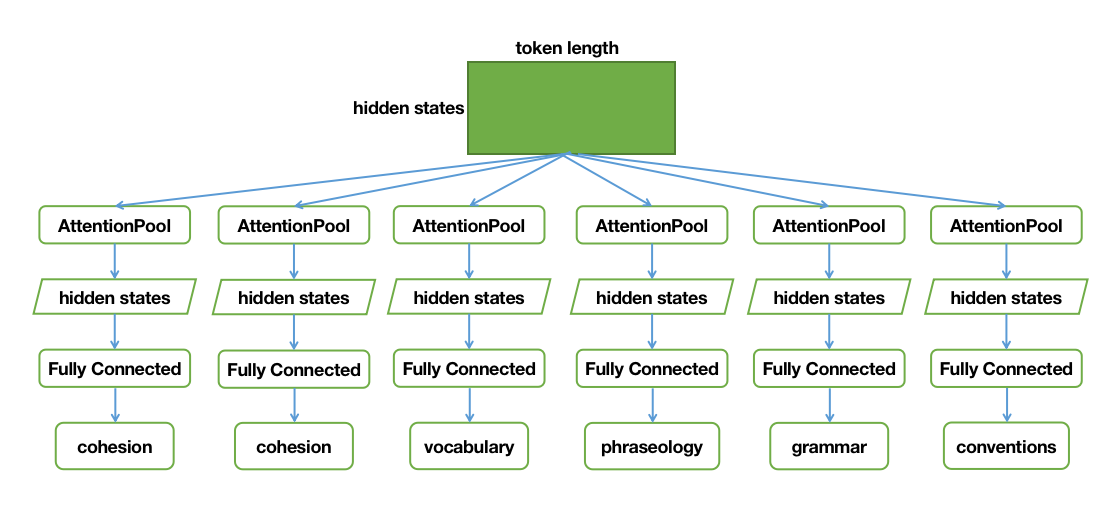}
            \end{subfigure}
            \caption{Illustrations of the 6 kinds of AttentionPooling (6AP) in our model}
            \label{fig:6AP_combined}
        \end{figure}

        Additionally, the AttentionPooling technique in DeBERTa can be represented as follows:

        \begin{equation}
        \text{AttentionPooling}(x) = \text{softmax}(\text{Pool}(x)) \cdot x
        \end{equation}
        
        where $x$ is the input sequence and $\text{Pool}(x)$ is a pooling function that captures the importance of each element in $x$.
        
        By leveraging DeBERTa with the 6 kinds of AP, we aim to develop more accurate and efficient automated feedback tools for ELLs in AES. These tools will provide targeted and personalized assessments, supporting the language development and academic success of ELLs.

            \subsection{Datasets}
        The dataset employed in this study, known as the ELLIPSE corpus, encompasses argumentative essays penned by English Language Learners (ELLs) in grades 8 to 12. These essays have undergone evaluation across six distinct analytical dimensions: cohesion, syntax, vocabulary, phraseology, grammar, and conventions.

        Each dimension signifies a specific facet of essay writing proficiency, wherein higher scores correspond to a heightened level of expertise in that particular dimension. Scores fall within the range of 1.0 to 5.0, with increments of 0.5. The primary objective of this competition involves predicting scores for all six dimensions in relation to the essays within the provided test set.
        
            \subsection{Evalution metrics}
        The evaluative criteria for this competition is the gracefully named 'Mean Columnwise Root Mean Squared Error' (MCRMSE), which is calculated as:

        \begin{equation}
        \textrm{MCRMSE} = \frac{1}{N_t}\sum_{j=1}^{N_t}\sqrt{\frac{1}{n} \sum_{i=1}^{n} (y_{ij} - \hat{y}_{ij})^2}
        \end{equation}
        
        where:
        \begin{itemize}
            \item $N_t$ is the number of columns,
            \item $n$ is the number of records per column,
            \item $y_{ij}$ is the true value,
            \item $\hat{y}_{ij}$ is the prediction.
        \end{itemize}

        The MCRMSE metric denotes the mean root mean squared error across all rows and columns. It gauges the precision of forecasts generated by the automated essay scoring (AES) models. The computation entails squaring the disparity between the actual value $y_{ij}$ and the projected value $\hat{y}_{ij}$ for each row and column, followed by averaging within each column, and ultimately taking the square root of the column averages. Lastly, the mean of these root mean squared errors spanning all columns is computed.

            \subsection{Results}

        The outcomes of the experiments, as presented in Table 1, illustrate the effectiveness of different backbone models and setups. For model evaluation, we adopted a 5-fold Cross-Validation approach employing MultiLabelStratifiedKFold to ensure an equitable allocation of labels. Model performance was assessed using the CV score, with lower scores indicating superior model effectiveness.

        \begin{table}[htbp]
        \centering
        \caption{Performance of Different Backbone Models and Configurations}
        \label{tab:results}
        \begin{tabular}{|c|c|c|c|}
        \hline
        \textbf{Backbone Model} & \textbf{CV Score} & \textbf{Pooling}        & \textbf{AWP} \\ \hline
        deepset/deberta-v3-large-squad2  & 0.45652 & AP & Yes \\ \hline
        deberta-large & 0.4542 & AP  & Yes \\ \hline
        microsoft/deberta-v2-xlarge & 0.4496 & AP & Yes  \\ \hline
        deberta-v3-large  & 0.4477 & 6 kinds of AP & No  \\ \hline
        deberta-v3-large  & 0.4473 & AP & Yes \\ \hline
        deberta-v3-large  & 0.4457 & 6 kinds of AP & Yes    \\ \hline
    
        \end{tabular}
        \end{table}

        Among the tested models, deberta-v3-large without Metric-specific AttentionPooling (6 kinds of AP) achieved a CV score of 0.4473. However, when AttentionPooling was enhanced with the Metric-specific AttentionPooling (6 kinds of AP) technique, the CV score decreased to 0.4457. This suggests that the incorporation of Metric-specific AttentionPooling (6 kinds of AP) improved the model's performance.

        Notably, the model with the lowest CV score (0.4457) and the highest performance utilized the deberta-v3-large backbone with Metric-specific AttentionPooling (6 kinds of AP) and Adversarial Weights Perturbation (AWP). While deberta-v3-large without Adversarial Weights Perturbation (AWP) achieved a CV score of 0.4477. This configuration demonstrates the effectiveness of Adversarial Weights Perturbation (AWP) in improving the performance of the automated essay scoring system.

        Furthermore, the deberta-large model with AttentionPooling achieved a CV score of 0.4542, while deepset/deberta-v3-large-squad2 and microsoft/deberta-v2-xlarge both achieved CV scores of 0.45652 and 0.4496, respectively, with AttentionPooling.
                
        Overall, the results indicate that incorporating Metric-specific AttentionPooling (6 kinds of AP) and Adversarial Weights Perturbation (AWP) techniques, specifically using the deberta-v3-large backbone, can lead to improved performance in automated essay scoring.
     
            \subsection{Conclusion}

        In conclusion, this paper has highlighted the advancements in automated essay scoring (AES) research and the need for addressing the specific requirements of English Language Learners (ELLs) in the context of writing proficiency assessment. The application of BERT-related techniques, such as DeBERTa, along with the incorporation of Metric-specific AttentionPooling (6 kinds of AP) and adversarial training through Adversarial Weights Perturbation (AWP), has shown promise in improving automated feedback tools for ELLs.

        The findings of this research contribute to the ongoing development of AES systems by providing insights into the effective assessment of ELLs' writing proficiency. The proposed solutions, including the use of DeBERTa, AWP, and 6 kinds of AP, offer a comprehensive approach to address the unique challenges faced by ELLs in language development. These advancements empower educators with automated feedback tools that are sensitive to the specific needs of ELLs, facilitating their language learning journey and overall academic success.

        Future research can further explore the optimization of hyperparameters and investigate the generalizability of these approaches to diverse populations of ELLs. Additionally, the integration of larger and more diverse datasets can enhance the robustness and inclusiveness of AES systems for ELLs. By continuously refining and advancing AES techniques, we can provide effective and tailored support to ELLs, fostering their language skills and educational achievements.

        \bibliographystyle{IEEEtran}
        \bibliography{references}
\end{CJK*}	
\end{document}